\documentclass[letterpaper]{article} 
\usepackage[draft]{aaai25}  
\usepackage{times}  
\usepackage{helvet}  
\usepackage{courier}  
\usepackage[hyphens]{url}  
\usepackage{graphicx} 
\usepackage{natbib}  
\usepackage{caption} 
\usepackage{algorithm}
\usepackage{algorithmic}
\usepackage{newfloat}
\usepackage{listings}

\usepackage{multirow}
\usepackage{pifont}       
\usepackage{bbding}       

\usepackage{amsmath}
\DeclareCaptionStyle{ruled}{labelfont=normalfont,labelsep=colon,strut=off} 
\floatstyle{ruled}
\newfloat{listing}{tb}{lst}{}
\floatname{listing}{Listing}
\title{Effectively Enhancing Vision Language Large Models\\ by Prompt Augmentation and Caption Utilization}

\usepackage{bibentry}

\begin{document}

\author{
	Minyi Zhao$^{1}$
 , Jie Wang$^2$, Zhaoyang Li$^2$, \\Jiyuan Zhang$^2$, Zhenbang Sun$^2$, Shuigeng Zhou$^1$\thanks{Corresponding author.}
 \\
 \small{\textmd{\textsuperscript{\rm 1}Shanghai Key Lab of Intelligent Information Processing,  and School of \\ Computer Science, Fudan University, Shanghai 200438, China\\
	$^2$ByteDance, China\\
 \{zhaomy20, sgzhou\}@fudan.edu.cn \\ \{wangjie.bernard, lizhaoyang.lee, zhangjiyuan, sunzhenbang\}@bytedance.com}}
}
 

\maketitle

\begin{abstract}
Recent studies have shown that Vision Language Large Models (VLLMs) may output content not relevant to the input images. This problem, called the hallucination phenomenon, undoubtedly degrades VLLM performance. Therefore, various anti-hallucination techniques have been proposed to make model output more reasonable and accurate. Despite their successes, from extensive tests we found that augmenting the prompt (\textit{e.g.} word appending, rewriting, and spell error etc.) may change model output and make the output hallucinate again. 
To cure this drawback, we propose a new instruct-tuning framework called \textbf{P}rompt \textbf{A}ugmentation and \textbf{C}aption \textbf{U}tilization (PACU) to boost VLLM's generation ability under the augmented prompt scenario. Concretely, on the one hand, PACU exploits existing LLMs to augment and evaluate diverse prompts automatically. The resulting high-quality prompts are utilized to enhance VLLM's ability to process different prompts. On the other hand, PACU exploits image captions to jointly work with image features as well as the prompts for response generation. When the visual feature is inaccurate, LLM can capture useful information from the image captions for response generation.
  Extensive experiments on hallucination evaluation and prompt-augmented datasets demonstrate that our PACU method can work well with existing schemes to effectively boost VLLM model performance. Code is available in \url{https://github.com/zhaominyiz/PACU}.
\end{abstract}

%
\section{Introduction}\label{sec:intro}
With the revolutionary progress of Large Language Models (LLMs)~\cite{raffel2020exploring,zheng2024judging,touvron2023llama,achiam2023gpt,jiang2023mistral} and various supporting techniques including pre-training~\cite{zoph2020rethinking,he2020momentum,chen2023vlp}, fine-tuning~\cite{zhang2018overview,jia2022visual,liu2022p,peng2023instruction}, parameter-efficient tuning~\cite{ding2023parameter} (\textit{e.g.} reparameterization~\cite{hu2021lora} and adapters~\cite{wang2020k} etc.), a large number of Vision Language Large Models (VLLMs)~\cite{li2022blip,li2023blip,liu2023improved,liu2024visual,zhu2023minigpt} have emerged. These VLLMs usually first use a feature extractor~\cite{dosovitskiy2020image} to obtain a visual feature from each input image. Then, this feature is processed and concatenated with the prompt to be fed into a LLM to generate the response. With the help of massive text-image resources~\cite{schuhmann2022laion}, rich instruct-tuning datasets~\cite{zhang2023instruction,yin2024lamm}, and powerful LLMs~\cite{liu2023summary}, these VLLMs have shown amazing potential in various computer vision tasks~\cite{fu2023mme} such as vision captioning~\cite{ghandi2023deep}, question answering~\cite{abdel2023deep}, and cross-modality retrieval~\cite{cao2022image}.

\begin{figure*}[t]
	\begin{center}
		\includegraphics[width=0.8\linewidth]{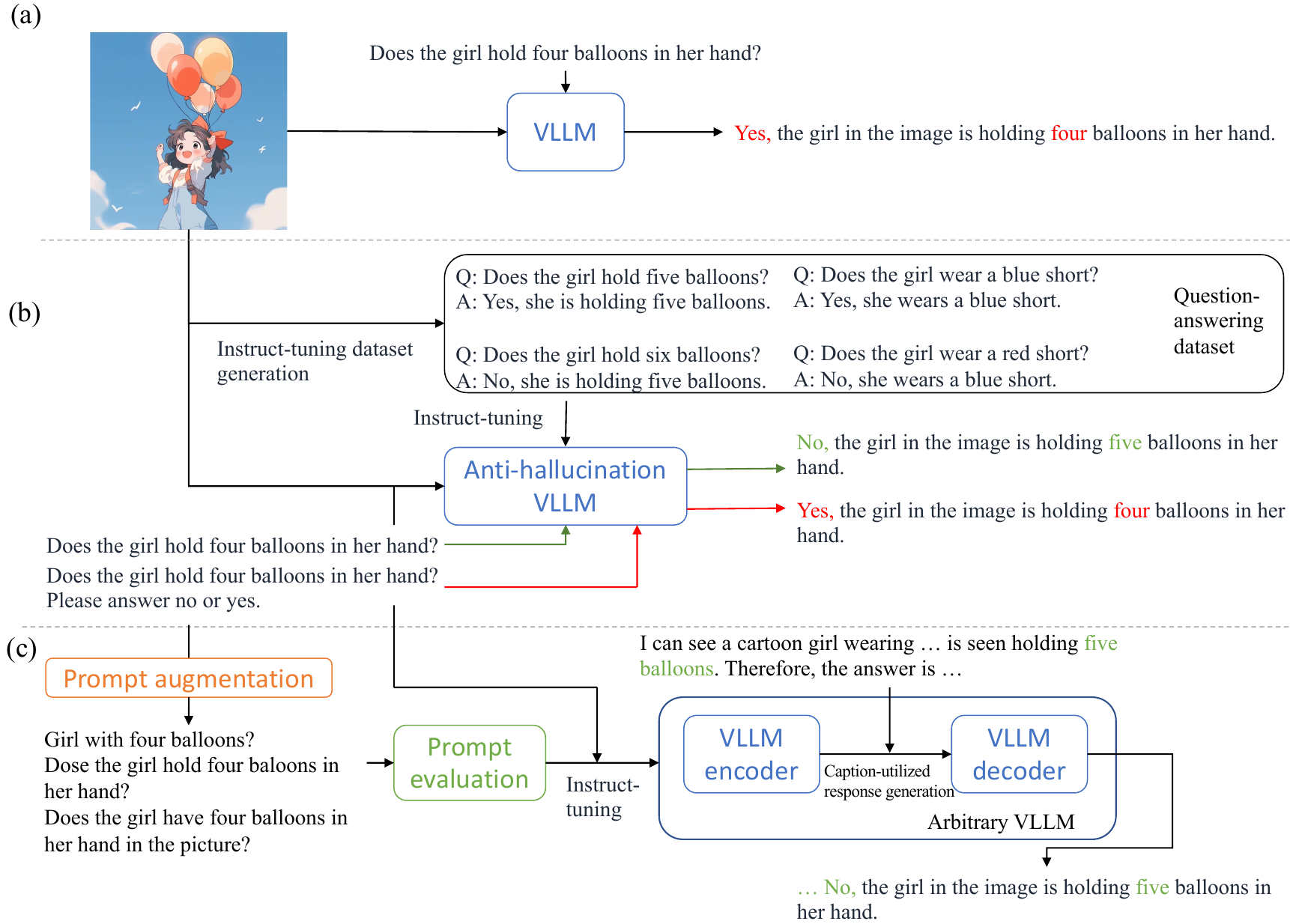}
	\end{center}
	\caption{Schematic illustration of existing works and our solution: (a) Generic VLLMs, (b) Anti-hallucination VLLMs, and (c) Our plug-and-play solution PACU that can boost existing VLLMs by prompt augmentation from the input side and caption utilization from the response generation side.}
	\label{fig:motivation}
\end{figure*}

Nevertheless, like LLMs, these advanced VLLMs also suffer from the hallucination problem~\cite{liu2024survey,zhou2023analyzing,wang2023evaluation}. As shown in Fig.~\ref{fig:motivation}~(a), when doing question answering (QA), these VLLMs may mistakenly count the number of balloons in the image, finally return a wrong response. Obviously, such a hallucination phenomenon severely impairs the performance and application of VLLMs. To address this drawback, various approaches are proposed to detect and overcome hallucinations via constructing QA pairs for instruct-tuning. In particular, POPE~\cite{li2023evaluating} extracts all the objects in each image to construct binary QA pairs. MME~\cite{fu2023mme} and CIEM~\cite{hu2023ciem} go further to cover more aspects of objects (\textit{e.g.} action, counting etc.) for more comprehensive VLLM performance evaluation. Yu \textit{et al.}~\cite{yu2023hallucidoctor} utilize multi-VLLM experts for better captioning.

As in LLMs, to further enhance VLLMs' prompt processing ability, we propose to introduce prompt augmentation~\cite{shum2023automatic}. 
However, from extensive tests
we observed that some prompt augmentation operations (\textit{e.g.} appending words and rewriting) may make VLLMs output hallucinate results. Taking Fig.~\ref{fig:motivation}~(b) for instance, if we append ``Please answer no or yes'' to the input prompt, the VLLM's response will turn back to a wrong answer. By further analyzing the tests, we found that the QA accuracy of existing VLLMs is significantly decreased --- taking InstructBLIP~\cite{dai2023instructblip} + Vicuna-1.1~\cite{zheng2024judging} for example, the accuracy on CIEM~\cite{hu2023ciem} dataset is degraded from 88.2\% to 79.5\%. What makes the matter worse is that such prompt augmentation has an apparent impact on response generation --- as shown in the 3rd prompt in Fig.~\ref{fig:motivation}~(b), by appending ``Two or three'', the response turns to a predesigned wrong answer. This indicates that typical VLLMs can not process augmented prompts well. Undoubtedly, this defect severely limits the performance and applications of VLLMs.


Then, \textit{how to boost VLLMs' augmented prompt processing ability?} To answer this question, 
this paper proposes a new general framework called PACU (the abbreviation of \textbf{P}rompt \textbf{A}ugmentation and \textbf{C}aption \textbf{U}tilization) to enhance existing VLLMs. Accordingly, as illustrated in Fig.~\ref{fig:motivation}~(c), PACU tries to boost existing VLLMs from both input and response generation sides simultaneously. 

From the input side, PACU generates various high-quality augmented prompts to enhance the model's prompt processing ability. To this end, an automatic prompt augmentation (PA) module is employed. PA consists of two sub-modules --- an LLM for prompt generation and an LLM used to evaluate the augmented prompts. Only high-quality prompts are retained and reweighted for model training. From the response generation side, PACU adopts a caption-utilized generation (CUG) mechanism to aid the response generation. As shown in Fig.~\ref{fig:motivation}~(c), when generating the response, VLLM is required to first briefly review the image and then use the caption as prior knowledge to assist the response generation so that when the visual feature is insufficient, LLM can still capture information from the caption, rather than only the prompt, to infer the response. Compared with typical approaches that usually use image caption as an instruct-tuning task for model training, we explicitly utilize captions as intuitive information to aid the response generation. The advantage of PACU is obvious: as a plug-and-play framework, PACU can collaborate with most existing anti-hallucination solutions, VLLM frameworks, and LLMs to further enhance VLLMs' prompt-processing capabilities.

Contributions of this paper are as follows: 1) We introduce prompt augmentation to VLLMs. And based on extensive tests, we find a severe drawback of existing VLLMs --- they cannot process augmented prompts well. 
2) We propose a new framework, PACU, to enhance VLLMs' abilities to process various prompts. On the one hand, PACU generates, evaluates and reweights various augmented prompts to develop the model. On the other hand, a caption-utilized generation mechanism is designed to generate the response with the help of image captions. 3) Extensive experiments on hallucination evaluation datasets demonstrate the effectiveness and advantage of our method when working with existing techniques to further lift VLLM performance.

\section{Related Work}

\subsubsection{Vision Language Large Models}
With the remarkable success of large language models (LLMs), many powerful models~\cite{guo2023images,li2023otter,bai2023qwen,li2023empowering,ye2023mplug}, which are called Vision Language Large Models (VLLMs), have been proposed to integrate LLMs with visual modality to do various visual language understanding tasks. Generally, these VLLMs first use an encoder (\textit{e.g.} ViT~\cite{dosovitskiy2020image}, EVA-ViT~\cite{fang2023eva}, and CLIP~\cite{radford2021learning}) to capture visual features from the input images. Then, various structures are utilized to align the visual features into the language domain. Specifically, a linear projection layer is applied by LLaVA~\cite{liu2024visual} and MiniGPT-4~\cite{zhu2023minigpt}. mPLUG-Owl~\cite{ye2023mplug} uses a visual abstractor to better align visual features. BLIP2~\cite{li2023blip} and InstructBLIP~\cite{dai2023instructblip} pre-train a Q-former to align and project visual features. Compared with BLIP2, InstructBLIP additionally fuses the prompt with the visual feature. Finally, the fused visual feature is concatenated with the prompt and fed into the LLM to obtain the response. Some works also use parameter-efficient tuning techniques (\textit{e.g.} LoRA~\cite{hu2021lora}) to fine-tune key layers of LLM to obtain better performance. The training processing of these VLLMs can also be divided into two stages --- pre-training stage and instruct-tuning. In the first stage, a large number of text-image pairs~\cite{schuhmann2022laion} are utilized to align the vision domain with the language domain. When it comes to the second stage, various instruct-tuning datasets are proposed to enable VLLMs to handle downstream tasks. For example, LLaVA~\cite{liu2024visual} feeds image captions to existing strong LLM to obtain QA pairs. Otter~\cite{li2023otter} collects a multilingual instruct-tuning dataset via visual annotation, strong LLM, and iterative mechanism. 

\subsubsection{Aniti-Hallucination Techniques}
Despite the fact that these advanced VLLMs above have achieved amazing progress in various downstream vision-language tasks, they still suffer from the hallucination phenomenon~\cite{liu2024survey} (\textit{e.g.} outputting non-existent objects and count error). To solve this problem, many efforts~\cite{gunjal2023detecting,li2023evaluating,hu2023ciem,fu2023mme,liao2023revo,liu2023aligning,wang2023vigc,zhou2023analyzing,yu2023hallucidoctor,wu2023v} have been paid to detect and then eliminate hallucination. These methods treat the hallucination phenomenon as inaccurate visual feature extraction and propose various techniques to lift the extraction. Among them, \cite{gunjal2023detecting} and POPE~\cite{li2023evaluating} consider the object hallucination and propose to detect the hallucination via a classifier and binary QA pairs, respectively. MME~\cite{fu2023mme} further extends the idea and proposes to evaluate the hallucination from 14 aspects. CIT~\cite{hu2023ciem} explores the idea of contrastive items, extends the idea to actions, and proposes to use chain-of-thought (COT) to supervise model training. Liu \textit{et al.}~\cite{liu2023mitigating} use GPT-4 to assist hallucination detection and construct more general QA pairs. VStar~\cite{wu2023v} proposes an iterative grounding technique to construct more fine-grained QA pairs. HalluciDoctor~\cite{yu2023hallucidoctor} proposes an automatic QA pair generation pipeline that consists of several VLLM experts, cross-checking, counterfactual
instruction synthesis mechanism to generate high-quality QA pairs. ShareGPT4V~\cite{chen2023sharegpt4v}, ALLaVA~\cite{chen2024allava}, and LLaVA-Next~\cite{liu2024llavanext} propose better caption techniques and higher-quality training data.
Although these techniques have successfully made VLLM output more accurate results, we find that after augmenting prompts as usual in LLMs for these VLLMs, their outputs will become hallucinate again. This indicates that further research is needed to boost VLLMs' diverse augmented prompt processing ability.

\subsubsection{Differences between Existing Works and Our Method}
The differences between existing anti-hallucination techniques and our PACU method are two-fold. On the one hand, the motivation is different, typical existing works mainly try to enable VLLMs to handle different tasks (\textit{e.g.,} cognition, recognition). Thereby, various instruct-tuning datasets are constructed to supervise the recognition of various objects and relationships in images. In contrast, the motivation of PACU is to boost VLLM's generation ability under diverse augmented prompts. Therefore, PACU first proposes to generate diverse prompts, then a caption-utilized generation mechanism is adopted to provide more information for response generation. On the other hand, the methodology is different, typical existing anti-hallucination works must generate various QA pairs. However, as a plug-and-play tool, PACU cooperates with existing anti-hallucination approaches to further boost model performance.

\begin{figure*}[t]
	\begin{center}
		\includegraphics[width=0.8\linewidth]{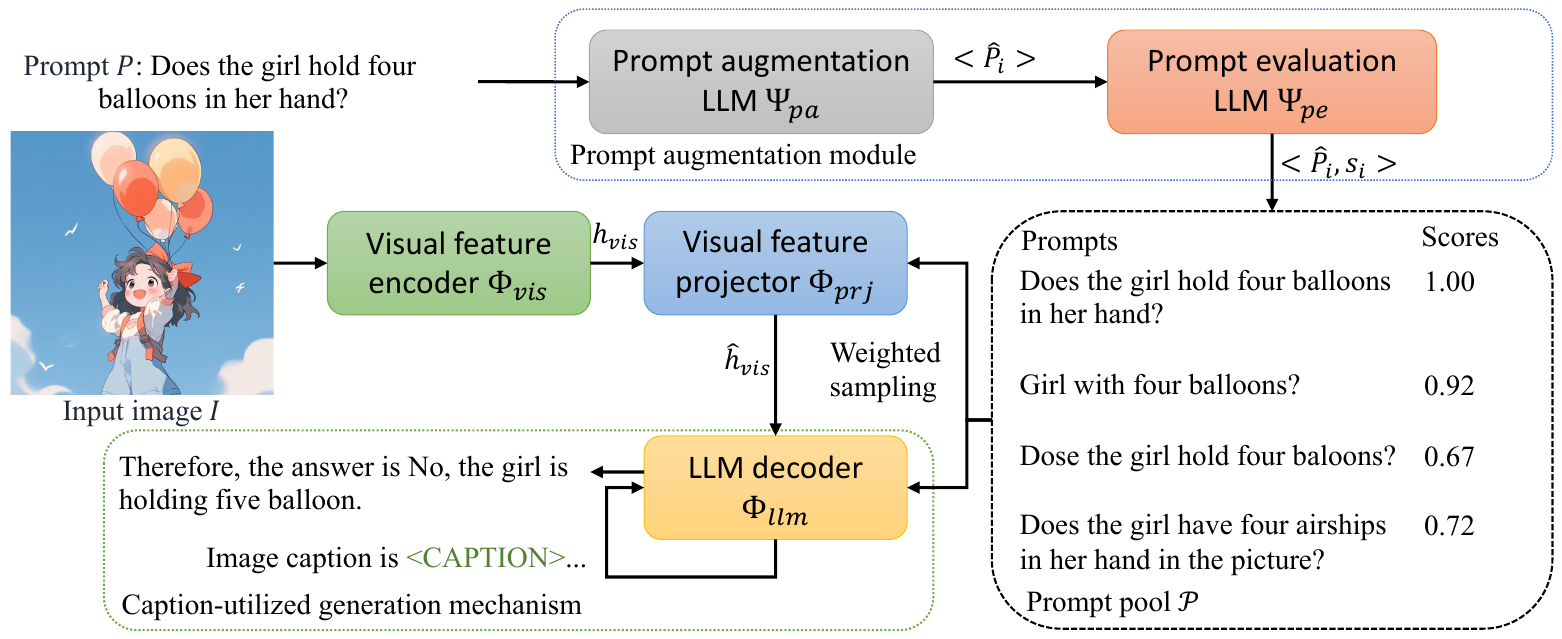}
	\end{center}
	\caption{The Framework of our PACU method.}
	\label{fig:pipeline}
\end{figure*}
\section{Methodology}
In this section, we present a detailed introduction to our proposed PACU method.

\subsection{Framework}
Fig.~\ref{fig:pipeline} is the framework of our PACU method. Unlike general VLLMs, PACU adopts a \textit{prompt augmentation} (PA) module that contains a \textit{prompt augmentation LLM} $\Psi_{pa}$ to augment prompts and a \textit{prompt evaluation LLM} $\Psi_{pe}$ for prompt quality scoring. PACU also uses a \textit{caption-utilized generation} (CUG) mechanism to provide image captions to aid response generation. PACU works sequentially in the following seven steps: 1) Use a pre-trained vsisual feature encoder $\Phi_{vis}$ to extract an initial visual feature $h_{vis}$ for each input image. 2) Apply the prompt augmentation LLM $\Psi_{pa}$ to generate augmented prompts $\mathcal{P}=\{\hat{P}_i\}$. 3) Utilize the prompt evaluation LLM $\Psi_{pe}$ to evaluate the quality score $s_i$ of each augmented prompt $\hat{P}_i$. 4) Sample a prompt from the prompt pool $\mathcal{P}$ based on the scores $\{s_i\}$: $P' \sim \mathcal{P}$. 5) Project and align the visual feature $\hat{f}_{vis}$ to the language domain via a visual feature projector $\Phi_{prj}$. 6) Concatenate $\hat{f}_{vis}$ and $P'$, and feed them to the LLM $\Phi_{llm}$. 7) Generate the image caption and then use the caption to assist the final response generation. Among all these steps, the 1st, 5th, and 6th steps are common processes in common VLLMs, implemented by their own algorithms, while the rest steps are new procedures introduced by our method.

\subsection{Prompt Augmentation}
The goal of the prompt augmentation module is to provide high-quality prompts for model training. To achieve this, we first use a prompt augmentation LLM $\Psi_{pa}$ to augment the initial prompt $P$: $\{\hat{P}_i\}=\Psi_{pa}(P, T)$, where $T$ is a policy template. Here, we consider 7 different policies, which are: 1) Hard: making the prompt harder to be understood; 2) Easy: making the prompt easier to be understood; 3) Short: shorten the length of the prompt; 4) Long: lengthen the prompt a bit; 5) Rewrite: rewrite the prompt by synonym replacement; 6) Spell: introduce spell errors (up to two) to the prompt. 7) Append: append some words at the beginning or end of the prompt. After this step, we obtain a pool $\mathcal{P}$ of $N$ prompts, \textit{i.e.}, $\mathcal{P}=\{\hat{P}_1,...,\hat{P}_i,...,\hat{P}_N\}$. Nevertheless, prompt augmentation may generate some low-quality prompts that have different semantics from the original one~\cite{zhao2022epida}. Ergo, we use a prompt evaluation LLM $\Psi_{pe}$ to check the semantic similarity between the augmented prompts and the raw prompt: 
\begin{equation}
    \label{eq:sim}
    s_i = \Psi_{pe}(P)\cdot \Psi_{pe}(\hat{P}_i),
\end{equation}
where `$\cdot$' denotes the dot product while $\Psi_{pe}(P)$ and $\Psi_{pe}(\hat{P}_i)$ are the corresponding hidden states of the raw prompt and the $i$-th augmented prompt extracted by the prompt evaluation LLM. We further set those $s_i$ whose value $<\epsilon$ to 0 to block the influence of some erroneous prompts. Finally, in our PACU method, we sample high-quality prompts from the pool $\mathcal{P}$ based on their scores.

\subsection{Caption-utilized Response Generation Mechanism}
General VLLMs use the visual feature $\hat{f}_{vis}$ and the prompt $P$ to drive the response generation. In PACU, we  additionally exploit the image caption $C$ to assist response generation so that when auto-regressively generating the response, LLM can use $C$ as prior knowledge to guarantee the quality of the generated response:
\begin{equation}
    \label{eq:gen}
    R_j=\begin{cases}\Phi_{llm}([\hat{h}_{vis}, P]|R_{t<j})\left( j < |C|\right) \\ \Phi_{llm}([\hat{h}_{vis}, P]|[C, R_{|C|<t<j}])\left( j \ge |C|\right) \end{cases},
\end{equation}
where $j$ denotes the $j$-th auto-regressive decoding and $|C|$ is the length of the caption $C$. In implementation, let the original ground truth response be $G$, we embed the caption $C$ to the beginning of $G$ to compose the new ground truth response $\hat{G}$. Then, during decoding, when $j<|C|$, LLM will first output the content $C$, and thus used in the following response generation procedure.

\subsection{Overall Loss Function}
In our method, the overall loss is the combination of the original loss and the augmented one:
\begin{equation}
    \label{eq:loss}
    \mathcal{L} = l(\Phi(I,P), \hat{G}) + \lambda E_{P' \sim \mathcal{P}} l(\Phi(I,P'), \hat{G}),
\end{equation}
where $l$ is the cross-entropy loss and $\lambda$ is utilized to balance these two loss functions.

\section{Performance Evaluation}
In this section, we first introduce the implementation details. Then, we combine PACU with state-of-the-art VLLM techniques to show its effectiveness and superiority. Finally, we perform extensive ablation studies and discussions to validate the design of our method.



\subsection{Implementation Details}
All experiments are conducted on 8 NVIDIA A100 GPUs with 80GB memory. The PyTorch version is 2.0. We use the officially recommended training hyper-parameters, optimizers, and learning rate schedulers to train the models. Both $\epsilon$ and $\lambda$ are set to 0.5 as in \cite{yu2023hallucidoctor}. $N$ is set to 7, which means for each kernel operation, one prompt is augmented. Prompt augmentation LLM $\Psi_{pa}$ and prompt evaluation 
 LLM $\Psi_{pe}$ are implemented by GPT-3.5 and SimCSE~\cite{gao2021simcse}, respectively. 

Since PACU should work with existing QA datasets, we introduce two QA-based hallucination datasets for model training, which are CIT~\cite{hu2023ciem} and POPE~\cite{li2023evaluating}, respectively. In PACU, when QA pairs from these datasets are sampled, PA and CUG will be operated. Because both CIT and POPE are built based on MS-COCO, we can directly sample image captions for CUG.

\begin{table*}[t]
\centering
\scalebox{0.9}{
\begin{tabular}{c | c | c || c | c | c | c || c | c | c | c }
\hline
\multirow{2}*{VLLM Framework} & \multirow{2}*{LLM}& \multirow{2}*{Method} & \multicolumn{4}{c||}{Original CIEM} & \multicolumn{4}{c}{Prompt augmented CIEM}\\
\cline{4-11}
~&~&~&  Accuracy & F1 & Precision & Recall &  Accuracy & F1 & Precision & Recall\\
\hline
\multirow{6}*{InstructBLIP} & \multirow{2}*{Vicuna-1.1} & CIT &88.2\% &0.876 &\textbf{0.891} &0.861 &79.5\% &0.784 &0.799 &0.770\\
~ & ~ & PACU &\textbf{88.7}\% &\textbf{0.884} &0.873&\textbf{0.896} &\textbf{84.4}\% &\textbf{0.840} &\textbf{0.832} &\textbf{0.848}\\
\cline{2-11}
~ & \multirow{2}*{Flan-T5-XL}&CIT &87.2\% &0.865 &0.879 &0.852&81.0\% &0.796 &0.826 &0.768\\
~ & ~& PACU &\textbf{88.2}\% &\textbf{0.878} &\textbf{0.882} &\textbf{0.873} &\textbf{82.7}\% &\textbf{0.818} &\textbf{0.828} &\textbf{0.809}\\
\cline{2-11}
~ & \multirow{2}*{Mistral-7B}&CIT &88.2\% &0.873 &\textbf{0.908} &0.841 &81.1\% &0.803 &0.811 &0.795\\
~ & ~& PACU &\textbf{88.9}\% &\textbf{0.884} &0.888 &\textbf{0.881} &\textbf{84.9}\% &\textbf{0.842} &\textbf{0.850} &\textbf{0.834}\\
\hline
\multirow{2}*{MiniGPT-4} & \multirow{2}*{Vicuna-1.1}&CIT &83.8\% &0.831 &0.834 &0.828 &74.3\% &0.721 &0.757 &0.688\\
~ & ~& PACU &\textbf{86.7}\% &\textbf{0.864} &\textbf{0.851} &\textbf{0.877} &\textbf{81.2}\% &\textbf{0.811} &\textbf{0.787} &\textbf{0.838}\\
\hline
\multirow{5}*{LLaVA} 
& \multirow{5}*{Mistral-7B}& LLaVA-1.5 &86.7\% &0.879 &0.832 &0.931 &81.2\% &0.825 &0.800 &0.853\\
~& ~& ShareGPT4V &89.3\% &0.900 &0.875 &0.927 &82.7\% &0.832 &0.839 &0.826\\
~& ~& LLaVA-Next &88.5\% &0.891 &0.874 &0.909 &84.0\% &0.847 &0.843 &0.851\\
~& ~& CIT &91.3\% &0.916 &\textbf{0.919} &0.913 &85.5\% &0.859 &0.866 &0.852\\
~ & ~& PACU &\textbf{91.8}\% &\textbf{0.922} &0.913 &\textbf{0.931} &\textbf{88.1}\% &\textbf{0.885} &\textbf{0.888} &\textbf{0.882}\\
\hline
\end{tabular}}
\caption{Performance improvement of PACU on CIEM dataset.}
\label{tab:ciem}
\end{table*}

\begin{table}[t]
\centering
\scalebox{0.85}{
\begin{tabular}{c || c | c | c}
\hline
\multirow{2}*{Evaluation datasets} & \multicolumn{3}{c}{LLaVA} \\
\cline{2-4}
~ & LLaVA-1.5 & CIT & PACU \\
\hline
VQAv2 &77.2\% & 78.0\% &\textbf{78.8}\% \\
GQA & \textbf{61.8}\% &61.5\% &61.7\% \\
VisWiz &\textbf{55.8}\% & 50.3\% &50.2\% \\
SciQA &70.8\% & 69.9\% &\textbf{73.1}\% \\
TextVQA &54.8\% & 52.9\% &\textbf{56.3}\% \\
POPE-All &\textbf{86.7}\% & 86.3\% &\textbf{86.7}\% \\
MME-Perception &1403.5 & 1444.2 &\textbf{1446.2} \\
MMBench-En &66.0\% & 65.6\% &\textbf{68.6}\% \\
SEED-Bench-Img &\textbf{67.0}\% & 66.3\% &66.5\% \\
LLaVA-Wild &\textbf{76.2} & 71.4 & 72.1 \\
MM-Vet &\textbf{30.1} & 28.8 &30.0 \\
CIEM-Org &86.7\% & 91.3\% &\textbf{91.8}\% \\
CIEM-Aug &81.2\% & 85.5\% &\textbf{88.1}\% \\
\hline
\#Wins &6 &0 &\textbf{8}\\
\hline
\end{tabular}}
\caption{Performance evaluation of PACU on more evaluation benchmarks. We use LLaVA as the framework and Mistral-7B as the LLM.}
\label{tab:more}
\end{table}

\subsection{Performance Improvement on SOTA Approaches}
\subsubsection{Results on the CIEM dataset:}
\label{sec:ciem}
Here, we report the model performance on both the raw prompts and the augmented prompts. To this end, we first need to generate a high-quality augmented testset. Particularly, we first sample QA pairs from CIEM~\cite{hu2023ciem} testset and utilize the PA module to generate high-quality and diverse prompts. After filtering, a prompt augmented testset that contains around 10K QA pairs is constructed. Furthermore, we crowdsource the quality of a split (200) of the testset and we find 96\% questions are qualified. Then, we check three VLLM frameworks (InstructBLIP~\cite{dai2023instructblip}, MiniGPT-4~\cite{zhu2023minigpt}, and LLaVA~\cite{liu2024visual}) and three VLLMs (Vicuna-1.1, Flan-T5-XL, and Mistral-7B). 
Following \cite{hu2023ciem}, we use CCS~\cite{li2022blip} to pre-train InstructBLIP and MiniGPT-4 and then fine-tune models using CIT and PACU, respectively. As for the LLaVA framework, following \cite{chen2023sharegpt4v,liu2024llavanext}, we use LLaVA-1.5 as the baseline and then install corresponding techniques to LLaVA-1.5 for a fair comparison. 

From Tab.~\ref{tab:ciem}, we can see that: 1) Typical VLLMs cannot process augmented prompts well. For example, the accuracy of the LLaVA-Next is impaired from 88.5\% to 84.0\% (a 4.5\% deterioration). Similar performance deterioration can be observed from other VLLMs. This indicates that there is an urgent need to lift VLLMs' ability of handling diverse prompts. 2) Our PACU method can boost VLLMs' performance in prompt augmentation scenarios. Specifically, when using the InstructBLIP framework, the VLLM performances with all LLMs increase 4.9\%, 1.7\%, and 3.8\% for Vicuna-1.1, Flan-T5-XL, and Mistral-7B, respectively. This indicates the effectiveness of our PACU method. 3) PACU is also strong enough to beat related baselines in the prompt-augmented setting. For instance, PACU outperforms LLaVA-1.5, ShareGPT4V, LLaVA-Next, and CIT when using LLaVA as the framework. 4) Our PACU method can also lift the model accuracy under the original unaugmented prompt setting. For instance, InstructBLIP+Vicuna-1.1 gains a performance increase of 0.5\% (from 88.2\% to 88.7\%). This further demonstrates the advantage of our PACU method. 5) Our PACU method also cooperates well with various VLLM frameworks and LLMs. As shown in Tab.~\ref{tab:ciem}, all combinations' model performance is boosted.  This helps to justify that PACU is a general technique that can support different VLLM frameworks and LLMs.
\begin{figure*}[t] 
	\begin{center}
		\includegraphics[width=0.9\linewidth]{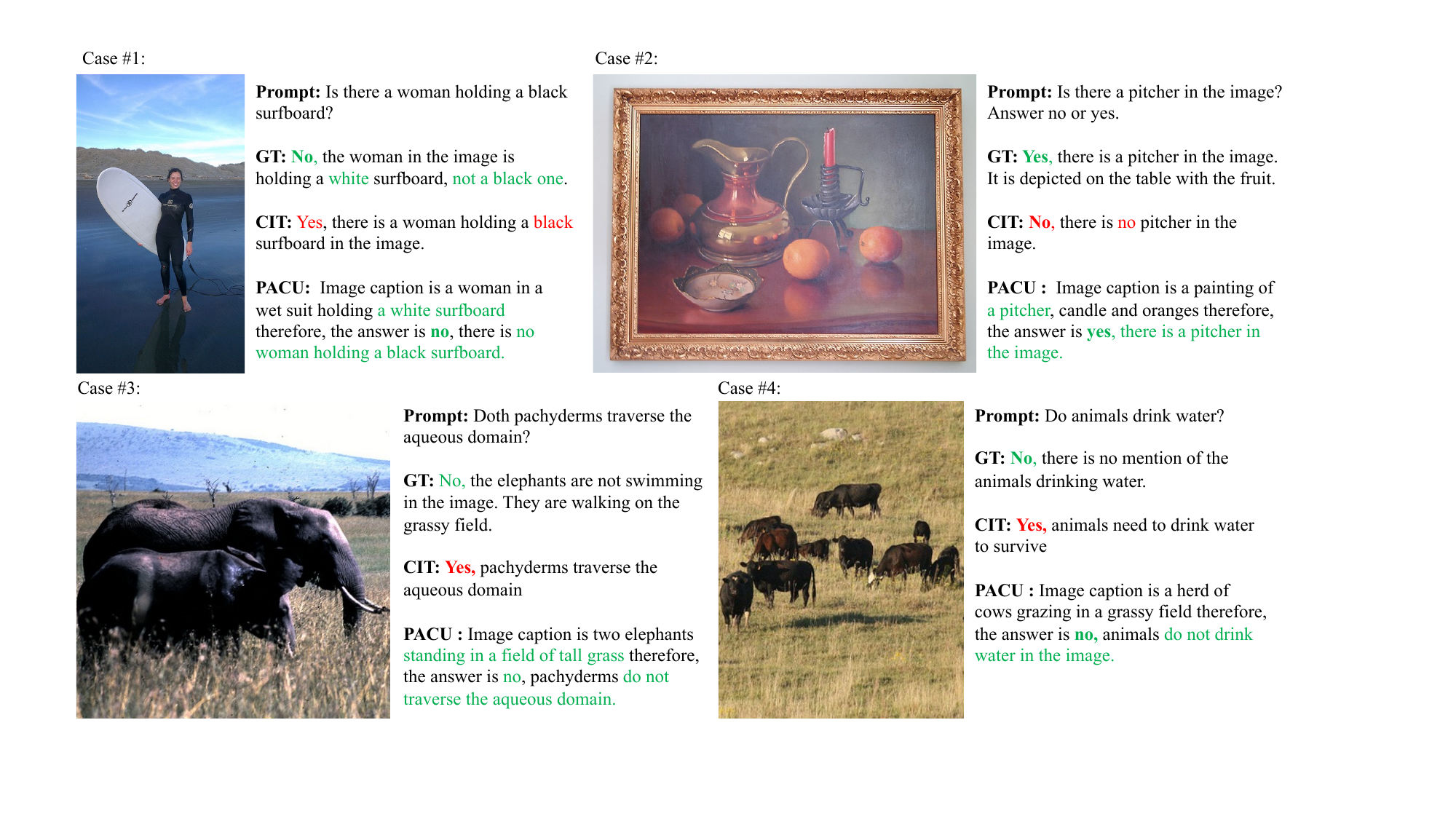}
	\end{center}
	\caption{Visual examples of generated responses of CIEM baseline and our PACU boosted model. We use InstructBLIP+Vicuna-1.1 to implement the VLLM. The incorrect part in the response is highlighted in red.}
	\label{fig:vis}
\end{figure*}
\subsubsection{Results on more evaluation datasets:} Here, we discuss the impact of PACU from the aspect of general capability. To this end, we develop an off-the-shelf VLLM for a wider model performance check. Accordingly, we consider the widely used training data -- LLaVA-1.5 and a total of 13 benchmarks, including VQAv2~\cite{goyal2017making}, GQA~\cite{hudson2019gqa}, VisWiz~\cite{gurari2018vizwiz}, SciQA~\cite{lu2022learn}, TextVQA~\cite{singh2019towards}, POPE~\cite{li2023evaluating}, MME~\cite{fu2023mme}, MMBench~\cite{liu2023mmbench}, SEED-Bench~\cite{li2023seed}, LLaVA-Wild~\cite{liu2024visual}, MM-Vet~\cite{yu2023mm}, Original CIEM (CIEM-Org)~\cite{hu2023ciem}, and Prompt-augmented CIEM (CIEM-Aug). We sample around 40K training data from CIT and then mix these data with LLaVA-1.5 as in \cite{chen2023sharegpt4v}. We utilize LLaVA as the VLLM framework and Mistral-7B as the LLM. All the experimental results are presented in Tab.~\ref{tab:more}.

From Tab.~\ref{tab:more} we can clearly see that after using PACU, the boosted model achieves better performance in more public evaluation datasets (8 \textit{v.s.} 6). This indicates that PACU will not impair the generality of VLLMs and is safe to use to faciliate VLLMs' augmented prompt processing abilities.

\begin{table*}[t]
\centering
\scalebox{0.8}{
\begin{tabular}{c | c | c || c | c}
\hline
Prompt augmentation & Prompt evaluation & Caption utilization & Acc of CIEM-Org & Acc of CIEM-Aug\\
\hline
\ding{51}& \ding{51}& \ding{51}& \textbf{88.7}\% &\textbf{84.4}\%\\
\ding{55}& \ding{55}& \ding{55}& 88.2\% &79.5\%\\
\hline
\ding{51}& \ding{55}& \ding{55}& 88.3\% &83.4\%\\
\ding{51}& \ding{51}& \ding{55}& 88.5\% &83.9\%\\
\ding{55}& \ding{55}& \ding{51}& 88.5\% &82.4\%\\
\hline
\end{tabular}}
\caption{Ablation study on the proposed prompt augmentation module and caption utilization mechanism.}
\label{tab:ab}
\end{table*}

\subsubsection{Visualization results:} Here, we provide some visualization cases to better demonstrate the advantage of our PACU method. From Fig.~\ref{fig:vis} we can see that: 1) Thanks to the caption utilization, the generated responses get better. Picking the 1st case for instance, CIT fails to recognize the color of the surfboard. However, with PACU, an accurate caption is generated to aid in inferring the correct color. 2) PACU also improves the model's ability to handle various prompts. In the 2nd, 3rd, and 4th cases, we feed prompts augmented by ``Append'', ``Hard'', and ``Short'' operations to the VLLM. We can see that the typical anti-hallucination method CIEM cannot cope with these augmented prompts. What is more, in the 4th case, LLM uses common knowledge, instead of visual features, to answer the question. After using PACU, VLLM can correctly answer these augmented prompts. These visualization results also demonstrate the superiority of our PACU method.

\subsection{Ablation Studies}
Here, we conduct extensive ablation studies to validate the
design of our method. We pick InstructBLIP~\cite{dai2023instructblip} and Vicuna-1.1~\cite{zheng2024judging} as the VLLM framework and the LLM, and report the model performance on both the original and augmented CIEM testset. All the experimental results are given in Tab.~\ref{tab:ab}.

\subsubsection{Overall PACU performance:} We first check the overall performance of PACU. From the 1st row and the 2nd row of Tab.~\ref{tab:ab}, we can see that PACU lifts the model performance from 88.2\%/79.5\% to 88.7\%/84.4\%, which demonstrates the impressive performance of our PACU method.

\subsubsection{The effect of prompt augmentation:} We design a variant model that introduces the prompt augmentation $\Psi_{pa}$ to the baseline (\textit{i.e.,} the 2nd row). From the 3rd row, we can see that prompt augmentation successfully advances the baseline from 88.2\%/79.5\% to 88.3\%/83.4\%. This indicates that doing prompt augmentation is beneficial for VLLM.

\subsubsection{The effect of prompt evaluation:} Here we check the effect of the prompt evaluation LLM $\Psi_{pe}$. We add $\Psi_{pe}$ to the aforementioned variant to check the model performance. As shown in the 3rd row and the 4th row, performing prompt evaluation can further lift the model performance (0.2\%/0.5\%). This justifies the benefit of improving training data quality through prompt evaluation.

\begin{table*}[t]

\centering
\scalebox{0.8}{
\begin{tabular}{c|c|c|c|c|c|c|c||c}
\hline
Method&Hard & Easy & Short & Long & Rewrite & Spell & Append&Overall\\
\hline
CIT&74.8\% & 85.0\% &82.7\% & 75.3\% & 79.8\% & 81.5\% & 74.4\% &79.5\%\\
PACU &\textbf{77.5}\% & \textbf{86.0}\% & \textbf{83.5}\% &\textbf{80.2}\% & \textbf{81.6}\% & \textbf{83.0}\% & \textbf{88.8}\% & \textbf{84.4}\%\\
\hline
\end{tabular}}
\caption{Performance degradation for each prompt augmentation operation.}
\label{tab:pa}
\end{table*}

\begin{table}[t]
\centering
\scalebox{0.8}{
\begin{tabular}{ c || c || c }
\hline
\multirow{2}*{Method} & CIEM-Org & CIEM-Aug\\
\cline{2-3}
~&  Accuracy &  Accuracy \\
\hline
POPE&67.5\%&66.1\%\\
PACU &\textbf{80.9}\%  &\textbf{70.2}\%\\
\hline
\end{tabular}}
\caption{Performance improvement of PACU working with POPE over the CIEM dataset.}
\label{tab:pope_ciem}
\end{table}

\subsubsection{The effect of caption utilization:} We finally check the effect of caption utilization. We also solely add the caption-utilized generation mechanism to the baseline model. Obviously, from the 5th row, we can see that the model performance of this variant (88.5\%/82.4\%) outperforms that of the baseline model (87.9\%/75.5\%). Thus, the advantage of the caption utilization is demonstrated.

\section{Discussions}
In this section, we discuss some important research issues. Unless otherwise specified, the implementation details in this section are the same as those in the ablation studies, \textit{i.e.}, InstructBLIP+Vicuna-1.1 and working with CIT.

\subsubsection{The performance of VLLM in resisting various augmented prompts:} As aforementioned, prompt augmentation will hurt typical VLLMs. Here, we report the detailed results to check the model's ability to handle different prompt augmentation operations. Tab.~\ref{tab:pa} presents the results. We can observe that: 1) As shown in the 1st row, for InstructBLIP+Vicuna-1.1, ``Hard'' and ``Append'' have the greatest impact on model performance, while ``Easy'' influences the least. 2) After applying our PACU method, model performance is significantly boosted on all the subsets. It is worth mentioning that on the ``Append'' subset, the performance improvement is the most significant. This is possibly because the ``Append'' operation does not cause significant changes to the sentence structure, so after fine-tuning, VLLM can easily handle it. 3) The performance on the ``Hard'' set is still low. This demonstrates the challenge of processing diverse prompts.

\subsubsection{The impact of different captions on PACU:} In PACU, captions are used to assist the response generation. Since CIT is built upon the MS-COCO dataset~\cite{lin2014microsoft}. We can directly sample captions from the annotations for CUG. Nevertheless, some datasets do not have caption annotations. Thereby, we use machine-generated captions (obtained by InstructBLIP+Vicuna-1.1) to assist the response generation. According to our experimental results, compared with PACU using human-labeled captions, using machine-generated captions leads to a slight performance deterioration -- from 88.7\%/84.4\% to 88.5\%/83.3\%. This indicates that the quality of captions will impact the caption-utilized mechanism. Considering that the performance of this variant is still significantly higher than the baseline model that has a corresponding performance of 88.2\%/79.5\%, and obtaining high-quality text-image data is not difficult, we believe that PACU is a promising and practical solution to enhance VLLMs.

\subsubsection{Alternative ways to assist LLM?} Here, we consider using tags to assist the final response generation. To achieve this, we first use RAM++~\cite{zhang2023recognize} to extract tags from images. Then, these tags are sorted via lexicographical order and are used to replace image captions. Unfortunately, we find that the model performance is degraded from 88.7\%/84.4\% to 83.5\%/78.6\%. It seems that tags are not suitable in PACU because the original pre-training does not include tagging. Therefore, more data training is required to use tags for assistance. In contrast, captioning has already been well pre-trained in VLLMs. Ergo, PACU uses image captions to assist response generation.

\subsubsection{How does PACU work with other anti-hallucination techniques?} Since PACU itself does not generate instruct-tuning data, PACU should work with existing anti-hallucination techniques. We have pinpointed the advantages of PACU work with CIT. To better check the universality of PACU, we also combine PACU with POPE. Experimental results are given in Tab.~\ref{tab:pope_ciem}. As shown in Tab.~\ref{tab:pope_ciem}, PACU significantly boosts the model performance from 67.5\%/66.1\% to 80.9\%/70.2\%. This indicates that PACU can also collaborate with other anti-hallucination works to improve model performance.

\subsubsection{Limitation and future work:} We have demonstrated the high performance and universality of PACU through extensive experiments. However, in PACU, VLLM is required to do captioning, which will slightly slow inference speed (on an A100 GPU, the inference cost of a batch deteriorates from 2.18s to 2.38s, 10\% slower). Besides, our proposed caption-utilized mechanism needs high-quality image captions to drive.
We leave the exploration of solutions of high performance and inference efficiency for future work.

\section{Conclusion}
In this paper, we study the prompt augmentation for VLLMs. In particular, we are the first to discuss the phenomenon that doing prompt augmentation may hallucinate existing VLLMs. Furthermore, to address this drawback, a new method called PACU is proposed. On the one hand, PACU introduces a prompt augmentation module to generate diverse prompts to enrich model's prompt processing ability. On the other hand, PACU exploits image captions to assist the response generation so that LLM can utilize useful information from the captions to finalize the response. Extensive experiments demonstrate the effectiveness, superiority and universality of our PACU method.
\section*{Acknowledgments}
The work was supported in part by a ByteDance Research Collaboration Project.

\bibliography{aaai25}
\end{document}